\documentclass[journal]{IEEEtran}

\IEEEoverridecommandlockouts                              

\usepackage{pgf} 

\usepackage{graphicx} 
\usepackage{times} 
\usepackage{amsmath} 
\usepackage{amssymb}  
\usepackage{changepage} 
\usepackage{longtable} 
\usepackage{algorithmic} 
\usepackage{bm}
\usepackage{gensymb} 
\usepackage{color}
\usepackage[belowskip=0pt]{caption}
\usepackage[hidelinks]{hyperref}
\usepackage[%
  vlined,
  boxed
]{algorithm2e}
\makeatletter
\renewcommand{\@algocf@capt@plain}{above}
\renewcommand{\algocf@caption@plain}{\box\algocf@capbox\vskip\AlCapSkip}%
\makeatother
\usepackage{tikz}
\usepackage{subcaption}

\SetKwProg{Fn}{Function}{}{end}\SetKwFunction{FRecurs}{FnRecursive}%

\title{\LARGE \bf
Deep Reinforcement Learning for Human-Like Driving Policies in Collision Avoidance Tasks of Self-Driving Cars
}

\author{Ran Emuna$^{1}$, Avinoam Borowsky$^{1}$, Armin Biess$^{1}$
\thanks{*This work was supported in part by the Helmsley Charitable Trust through the Agricultural, Biological and Cognitive Robotics Initiative and the Israel Science Foundation (grant no. 1627/17).}
\thanks{$^{1}$Department of Industrial Engineering and Management
Ben-Gurion University of the Negev
        {\tt\small}%
        }
}

\usepackage{marginnote}
\usepackage{soul}

\begin{document}

\newcommand{\bea}{\begin{eqnarray}}
\newcommand{\eea}{\end{eqnarray}}
\newcommand{\nn}{\nonumber}
\newcommand{\pa}{\partial}

\tikzset{
	cframe/.pic={
		\draw [->, thick, red] (0, 0) -- (3, 0);
		\draw [->, thick, green] (0, 0) -- (0, 3);
	}
}
\tikzstyle{box}=[rectangle, draw=black, rounded corners, text centered, anchor=north, minimum height=1.5cm, text width=3cm, thick]

\maketitle
\thispagestyle{empty}
\pagestyle{empty}

\begin{abstract}
The technological and scientific challenges involved in the development of autonomous vehicles (AVs) are currently of primary interest for many automobile companies and research labs. However, human-controlled vehicles are likely to remain on the roads for several decades to come and may share with AVs the traffic environments of the future. In such mixed environments, AVs should deploy human-like driving policies and negotiation skills to enable smooth traffic flow. To generate automated human-like driving policies, we introduce a model-free, deep reinforcement learning approach to imitate an experienced human driver's behavior. We study a static obstacle avoidance task on a two-lane highway road in simulation (Unity). Our control algorithm receives a stochastic feedback signal from two sources: a model-driven part, encoding simple driving rules, such as lane-keeping and speed control, and a stochastic, data-driven part, incorporating human expert knowledge from driving data. To assess the similarity between machine and human driving, we model distributions of track position and speed as Gaussian processes. We demonstrate that our approach leads to human-like driving policies.

\end{abstract}

\section{INTRODUCTION}

Self-driving cars are an emerging technology that may have major impact on our society, economy and environment in the 21st century. However, many legal, technical and algorithmic challenges have yet to be resolved in order to enable  fully autonomous vehicles (AVs). For example, the incorporation of prior knowledge, better understanding of the car surrounding's context (e.g., traffic signs, pedestrians, obstacles, vehicles) as well as the decision-making process resulting in driving policies for different environments and scenarios, are at the focus of current industrial  and academic research \cite{shalev2017formal}. Due to these ongoing research efforts, one possible scenario for the deployment of AVs may be that conventional vehicles and vehicles with (steadily) increasing autonomy will share roads for some period of time in the future. These mixed traffic environments will impose constraints on possible AV driving policies in order to maintain efficient and safe traffic flow. We argue here that autonomous vehicle should exhibit human or near human-like driving behavior. First, human drivers need to anticipate and predict other drivers' behavior because any unexpected behavior may induce conflicts between drivers and pose potential threats. Second, machine driving in AVs may be efficient, for example, in terms of fuel consumption or time, but driving behaviors, such as sudden stops, reckless or overcautious driving, may induce discomfort among human passengers. Thus, machine driving algorithms need to consider human factors.
\begin{figure}[t]
	\centering
	\vspace*{0.1cm}
	\begin{subfigure}[b]{0.49\textwidth}
		\begin{tikzpicture}
		\node[anchor=south west,inner sep=0] at (0,0) {\includegraphics[width=\textwidth, clip=true]{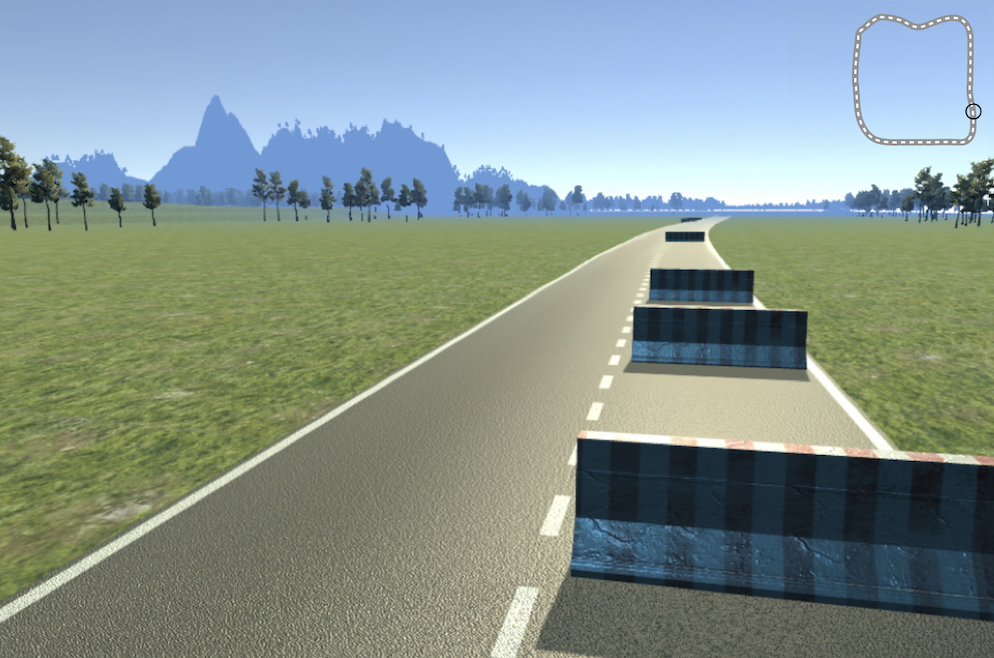}};
		\end{tikzpicture}
		\caption*{} \label{fig:simulator_env:a}
	\end{subfigure}
	\caption[]{Simulation environment for a static obstacle avoidance task in Unity. The environment consists of a two-lane highway road (4.7 km) with static obstacles placed along the road. The insert shows the overall geometry of the road track. The  circle indicates the location of the snapshot.}
	\label{fig:simulator_env}
\end{figure}

Programming an AV is not a trivial task. Traditionally, a complete AV system consists of different, interconnected  engineering stacks or layers. In a nutshell, sensory and map inputs are used for detection, tracking, prediction, planning and control \cite{urmson2008autonomous}. In the last decade, these stacks have been augmented by machine learning methods resulting in a combination of data-driven and rule-based methods \cite{fridman2019advanced}.

Much progress has been made since the first neural network-driven car was introduced in 1988 \cite{pomerleau1989alvinn} and the execution of three DARPA autonomous car challenges  in 2004, 2005, \cite{thrun2006stanley}, and 2007, \cite{urmson2008autonomous, buehler2009darpa}. By 2019, for example, Waymo reported to have reached 10 million miles of fully autonomous driving (level 4) \cite{chishiro2019towards}, whereas Tesla announced to have reached 1 billion miles with its semi-autonomous autopilot (level 2) \cite{fridman2019tesla}. Despite these impressive achievements, there is no large scale deployment of AVs yet. 
Many environments and traffic scenarios are still too challenging for an AV to handle, in contrast, to a skilled human driver. One of the main advantages of a human driver is its ability to anticipate and handle unfamiliar and novel situations. Human drivers can rely on prior knowledge and memory structures known as schemata \cite{borowsky2008relation}. A memory schema may be regarded as an abstract representation of the current situation based on previous encounters with similar situations. These structures allow drivers to perceive driving relevant elements even if these elements are not currently present in the scene or are obscured by the road environment. These schemata can help drivers in the selection of responses, which are appropriate for the current driving situation. One central question is therefore how can a machine learn from human drivers and develop human-like driving skills?

Deep neural networks (DNNs) are currently a major tool for artificial intelligence and have been used to learn from large amounts of data with minimal human intervention \cite{goodfellow2016deep}. Deep reinforcement learning (DRL), i.e., the combination of deep learning with reinforcement learning (RL), has been in particular successful in generating agents that can learn and act in uncertain, large and stochastic environments, see \cite{li2017deep}, \cite{ franccois2018introduction},  for a review.

In this paper, we apply a DRL approach to a static obstacle avoidance task and study how a machine-driven agent can imitate human driving behavior. A reinforcement learning algorithm provides a natural framework to combine data-driven methods--to incorporate human expert driving --with rule-based methods -- to encode basic driving rules.  We aim to imitate human driving behavior by recovering the first and second moments of state distributions of an experienced human driver. Unlike other imitation learning approaches, such as inverse reinforcement learning \cite{abbeel2004apprenticeship}, the reward function is constructed explicitly from basic driving rules and human driving data. Similar to \textit{novice} human drivers, we assume that the learning agent  has no exact knowledge about the underlying dynamics. As early studies in modeling of human driving behavior have emphasized, a human driver and a vehicle need to be considered as one \textit{combined} system, thus, making a dynamical system description  a challenging task as mechanical  \textit{and} human factors need to be considered \cite{macadam2003understanding}. In addition, a model-free approach is general and often easier to implement.

\section{RELATED WORK}

Modeling of human driving behavior can be roughly divided into two major approaches: model/rule-based  and data/learning-based methods. Model-based methods require prior knowledge, which may be obtained from experiments, physics, and/or the cognitive sciences and are often divided into perception, decision, and execution modules.  For reviews, we refer to \cite{ranney1994models}, \cite{macadam2003understanding}, \cite{fuller2005towards}, \cite{plochl2007driver}.

Recent rule-based methods for basic driving tasks in autonomous vehicles, such as steering, speed control, overtaking and obstacle avoidance, include model predictive control, 
 \cite{kong2015kinematic, qu2014switching}, Markov chain models \cite{althoff2009model, zou2016real}, and adaptive control \cite{petrov2014modeling, koh2015integrated}. These methods usually do not aim to imitate human driving behavior directly.

In the last years, data-driven methods, in particular, deep neural networks, often in combination with reinforcement and imitation learning,  have been used to model various driving tasks in different environments \cite{grigorescu2019survey}. 

Reinforcement learning methods are based on a reward function and are often not designed to reproduce human-like behavior. For example, Lillicrap et al,  \cite{lillicrap2015continuous}, have used a deep deterministic policy gradient (DDPG) RL algorithm to control a car in the simulator environment TORCS \cite{loiacono2013simulated}, in both, low-dimensional (sensory data) and high-dimensional (pixels) state spaces. At each step, a positive reward was provided for moving forward and a penalty of $-1$ for collisions. While this work successfully exemplified a highway driving control scenario, it did not aim for human-like driving. In contrast, a combination of offline supervised learning and online RL learning methods using human driving data from a simulator have been applied in \cite {lu2018learning} to imitate mean human behavior in overtaking tasks. This work is in its objectives similar to our study. Here, we aim to recover also the variability in human driving.

Imitation learning methods  are by default aiming to reproduce human expert behavior, for an excellent review see \cite{osa2018algorithmic}. Two major approaches are used: behavioral cloning (BC), \cite{pomerleau1989alvinn}, and inverse reinforcement learning (IRL), \cite{ng2000algorithms, abbeel2004apprenticeship}.
BC-based imitation learning methods for driving  have been used in \cite{bojarski2016end} in an end-to-end approach using a convolutional neural network (CNN). Large scale crowd-sourced video data have been used in Xu et al, \cite{xu2017end}, to learn  visuomotor action policies from current visual observations and previous vehicle states using long short-term memory (LSTMs) recurrent neural networks. However, one drawback of these methods consists in the inability of the learner to exceed the performance of the expert. To overcome these limitations additional elements have been incorporated. For example, Codevilla et al, \cite{codevilla2018end}, have implemented an end-to-end imitation learning approach using CNNs, in which input of the expert's intention at test time can be used to resolve the ambiguity in the perceptuomotor mapping.
Chen at al, \cite{chen2015deepdriving}, have used a CNN for mapping input images to affordance indicators (distance to lane markings and preceding cars, heading angle), which are then mapped to driving controls using a model. 
Finally, Levine and Koltin, \cite{levine2012continuous},  have implemented an IRL algorithm using human demonstrations
of simulated driving tasks and used tasked-relevant statistics to compare human and machine driving behavior. A similar approach is chosen here to compare human and machine driving.

Our contributions are as follows. First, we provide a simulation environment to study human and machine driving and formulate a model-free reinforcement learning framework to imitate the behavior (mean and variability) of an experienced human driver (the expert) by combining  data-driven with rule-based methods for reward engineering. Second, we use an existing state-of-the-art algorithm (PPO) in combination with a Mixture Density Network (MDN) to produce more flexible stochastic policies and introduce dynamic batch update for more efficient learning. Third, we model human and machine driving by Gaussian processes (GPs) and assess similarity of driving behavior between human and machine by comparing resulting distributions. Finally, we test generalization capabilities of the agent on  new roads with different obstacle distributions.

\section{METHODS}

\subsection{Simulation Environment}
\label{sec:simulator_env}
For this study we have built a flexible driving-simulator environment  in Unity, a professional game engine written in C\#. Unity gives the ability to create road environments and complex traffic scenarios.  A \textit{Logitech G29 Driving Force} steering wheel and pedals were used  to manually control the car and to collect data from experienced drivers, as will be described in Section \ref{sec:data_acquisition}.  An interface for Python was implemented in which all algorithms were written.  All our simulations were performed on a PC (i7 3.7 GHz, 32 GB RAM, GeForce GTX 1080 Ti).

\textit{Road Structure and Use Case:}
\label{sec:road_structure_use_case}
The road consisted of a closed track with straight and curved sections of approximately 4.7 km length and two lanes, 6 meters width each (see insert, Figure \ref{fig:simulator_env}).

\noindent
An obstacle avoidance task was designed by placing seventeen obstacles (barricades) along the road, nine on the right lane and eight on the left lane, as shown in Figure \ref{fig:simulator_env}.

\textit{State and Action Space}: Table \ref{table:state_and_action} presents the state variables of the environment and the action variables of the agent. 

\begin{figure}[ht]
    \centering
    \includegraphics[width=0.5\textwidth]{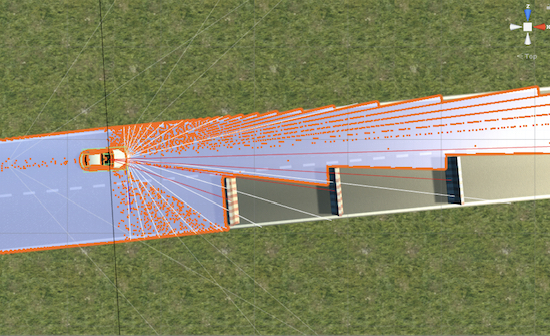}
    \caption{Range sensor measurements:  The range sensor, providing the observation measurement $ob$, has a field of view of $360\degree$ and a maximal range of $300$ meters. The field of view is covered by 288 equally separated rays (in steps of $1.25 \degree$). Each ray determines the distance from the vehicle to the nearest road obstacle or road boundary.}
    \label{fig:ob_sensor}
\end{figure}

The state-space has 310 dimensions and is defined by two subsequent observations (except the range sensor (rs) measurements $ob$) to infer temporal information. The latter significantly improved the performance of the algorithm and training time. When exceeding road boundaries, the values of the range sensor are not reliable and are set to $-1$. A visualization of the range sensor is shown in Figure \ref{fig:ob_sensor}.

The maximum speed of the vehicle was set to 100 km/h.  The  components of the state-vector have different physical units (for example, speed versus heading). To facilitate learning, we scaled all variables between $(-1, 1)$ and $(0,1)$, respectively, depending on whether the variable can take negative values or not. The action space has two dimensions, consisting of steering angle and braking/acceleration commands. Negative values represent braking, whereas positive values indicate acceleration input.

\begin{table*}[h]
\caption{State and action spaces.}
\begin{center}
\begin{tabular}{ |c||l|c|c|  }
\hline
\multicolumn{4}{|c|}{State Variables} \\
\hline
Notation & Description & Range & Scale\\
\hline
$\psi$       & steering & $[-25\degree, 25\degree]$ & $[-1,1]$ \\
$\tau$       & torque: brake + gas & $[-1,1]$ & $[-1,1]$ \\
$V$          & speed & $[0, 100]$\,km/h & $[0,1]$ \\
$\theta$     & heading & $[-\pi, \pi]$\,rad & $[-1,1]$ \\
$D$          & distance to the center line & $[-6,6]$\,m & $[-1,1]$ \\
$D_r$        & distance to the right lane & $[-9,3]$\,m & $[-1,\frac{1}{3}]$ \\
$D_l$        & distance to the left lane & $[-3,9]$\,m & $[-\frac{1}{3},1]$ \\
$\delta_{fr}$  & nearest front obstacle in the right lane & $[0,300]$\,m & $[0,1]$ \\
$\delta_{fl}$  & nearest front obstacle in the left lane & $[0,300]$\,m & $[0,1]$ \\
$\delta_{br}$  & nearest back obstacle in the right lane & $[0,300]$\,m & $[0,1]$ \\
$\delta_{bl}$  & nearest back obstacle in the left lane & $[0,300]$\,m & $[0,1]$ \\
\textit{rs}  & 288 obstacles and road boundaries sensors & $[0,300]$\,m & $[0,1]$ \\
\hline
\multicolumn{4}{|c|}{Action Variables} \\
\hline
$\psi$       & steering & $[-25\degree, 25\degree]$ & $[-1,1]$ \\
$\tau$       & torque: brake + gas & $[-1,1]$ & $[-1,1]$ \\
\hline
\end{tabular}
\label{table:state_and_action}
\end{center}
\end{table*}
\subsection{Human Driving Data Acquisition}
\label{sec:data_acquisition}
To collect data from experienced drivers, we performed an experiment with four participants. Each of the participants repeatedly traveled the road for eight rounds, while providing acceleration/braking and steering controls. The participants were instructed to drive  normally  as they would do under similar real-world conditions with the following additional instructions: (1) keep driving on the right lane when possible (2) keep the speed limit of 100 km/h. Each participant drove two test rounds to get familiar with the simulation environment. During the experiment, the state-vector was recorded with a sampling rate of 10 Hz, together with the global world coordinates of the car position. The global world coordinates were used to calculate the traveled distance from the start position by accumulating the Euclidean distance of subsequent car locations. Note that the latter corresponds to odometer measurements in the car. During the experiment, we noticed that two of the participants had problems to keep the car in the lane. We therefore selected one participant, which showed more stable behavior, as the human expert driver.

\subsection{Modeling Human Driving Behavior}
\label{sec:modeling_human_behavior}
Modeling human driving behavior is a difficult task since many factors are affecting driving, such as traffic environments, traffic participants, traffic rules, driver intentions and internal driver states, just to name a few \cite{leung2016distributional}. Generally, driving is described by a dynamic, stochastic and partially observable environment, and thus, poses big challenges to any modeling approach. A large amount of literature exists for modeling various aspects of human driving \cite{borrelli2005mpc, kong2015kinematic}. 
In a reinforcement learning framework -- which we  apply here --  a  reward function needs to be engineered to model human-like driving. To encode all aspects of driving this reward fucntion has presumably a rather complicated form. To facilitate the  process of reward shaping we decompose the reward function into two parts: a model-driven part encoding basic driving rules, such as staying within the road boundaries, speed limits and smoothness assumptions, and a data-driven part, which incorporates expert knowledge using driving data.

\begin{figure}[ht]
    \centering
    \begin{minipage}[b]{0.5\textwidth}
    \includegraphics[width=\textwidth]{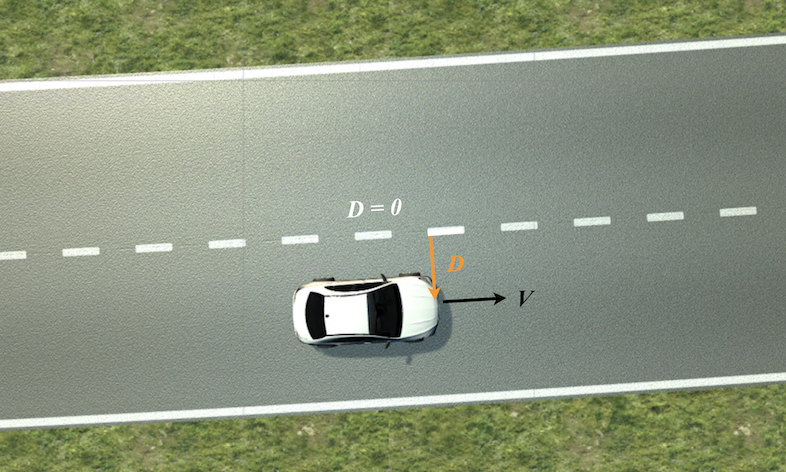}
    \end{minipage}
    \caption{Illustration of track position ($D$). The track position is measured with respect to the road's center-line (dashed line, $D=0$), which divides the road into two lanes. The track position takes values between the left (-6m) and  right lane boundary (6m), whereas the left and right center-line (not visible) is located at -3m and 3m, respectively.}
    \label{fig:trackPos}
\end{figure}

Expert driving is represented by two stochastic variables, track-position ($D$) and speed ($V$). The corresponding distributions are modeled using Gaussian processes (GPs), where the traveled Euclidean distance (arc-length) is the independent variable. Parametrization with respect to arc-length -- rather than time -- allows to compare trajectories between human and machine. Track-position $D$ refers to the signed distance of the car from the road center-line, as illustrated in Figure \ref{fig:trackPos}.  For constructing the GPs, a \textit{rational quadratic kernel} is chosen  and the noise variance is tuned such that the demonstrated trajectories were covered inside a 99\% confidence interval. The parameters of the GPs were optimally adapted to the data by maximizing the log-marginal-likelihood \cite{murphy2012machine}.

\begin{figure}[ht]
    \centering
    \begin{minipage}[b]{0.49\textwidth}
     \includegraphics[width=\columnwidth]{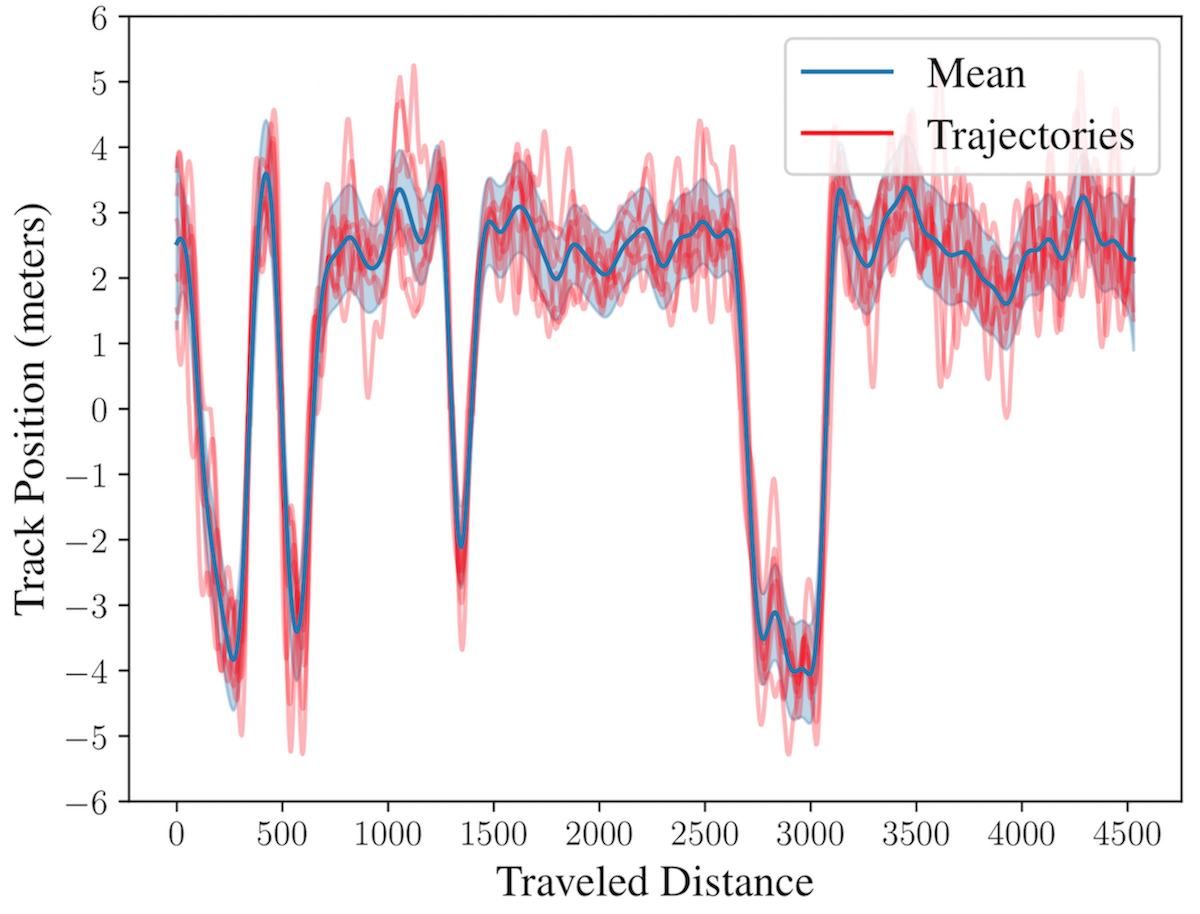}
    \caption*{(a)}
    \end{minipage}
    \vfill
    \begin{minipage}[b]{0.49\textwidth}
     \includegraphics[width=\columnwidth]{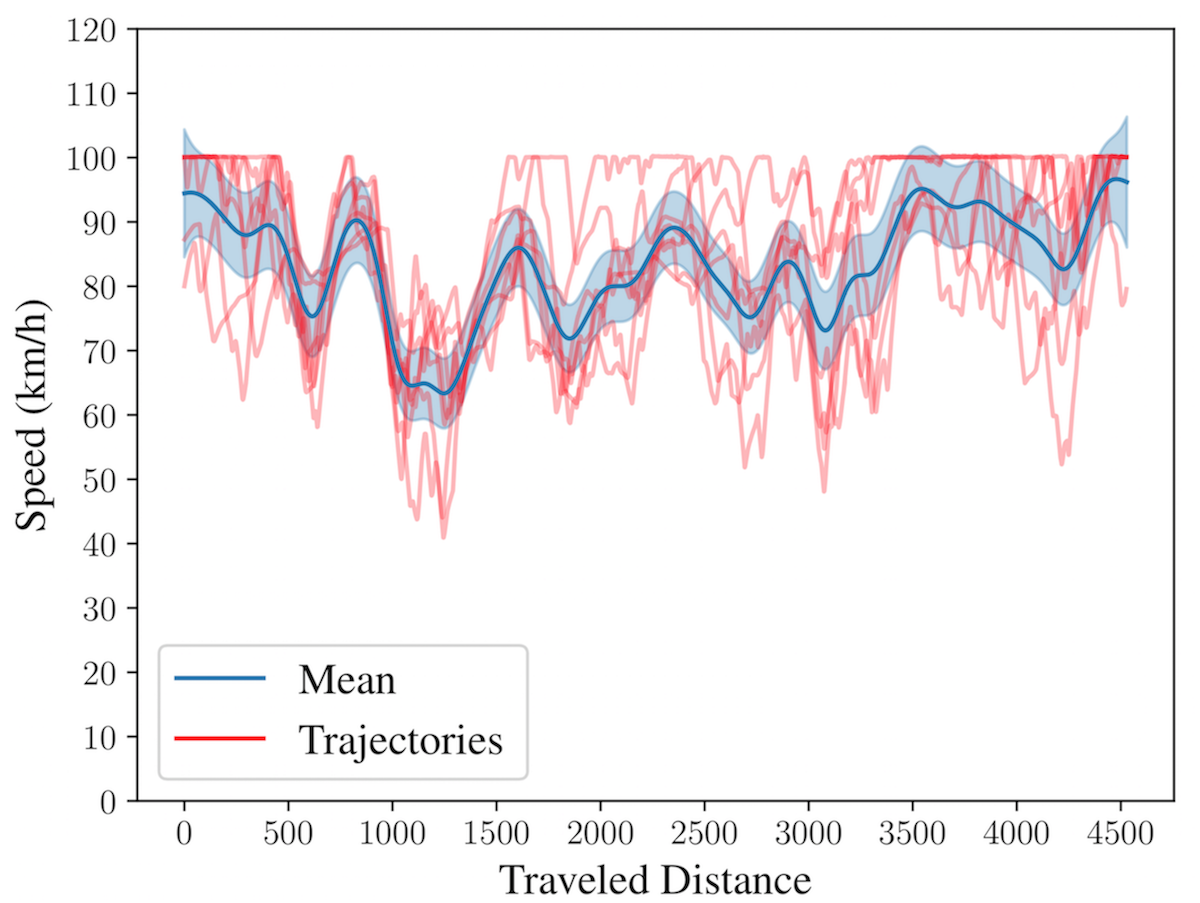}
    \caption*{(b)}
    \end{minipage}
    \caption{Human expert driving and GP regression: trajectories (red) and GPs (blue) of (a) track position and (b) speed as a function of traveled distance. The blue line shows the mean and the shaded blue area represents a 99\% confidence interval.}
    \label{fig:GP_trackPos}
\end{figure}

Figure \ref{fig:GP_trackPos}a and \ref{fig:GP_trackPos}b show the measured distribution for track position and speed of the expert driver, respectively,  as well  as the corresponding modeled distributions  using GPs. After modeling the expert driver's probability distribution  of the two measurement variables ($D$ and $V$), we  sampled 100 trajectories off-line (per measurement) within a bound of a 99\% confidence interval. These samples are then used in the evaluation of the  reward function as shown in the next section. A sample size of 100 was chosen  to cover well a 99\% confidence interval.
\begin{figure}[ht]
    \centering
    \begin{minipage}[b]{0.5\textwidth}
    \includegraphics[width=\columnwidth]{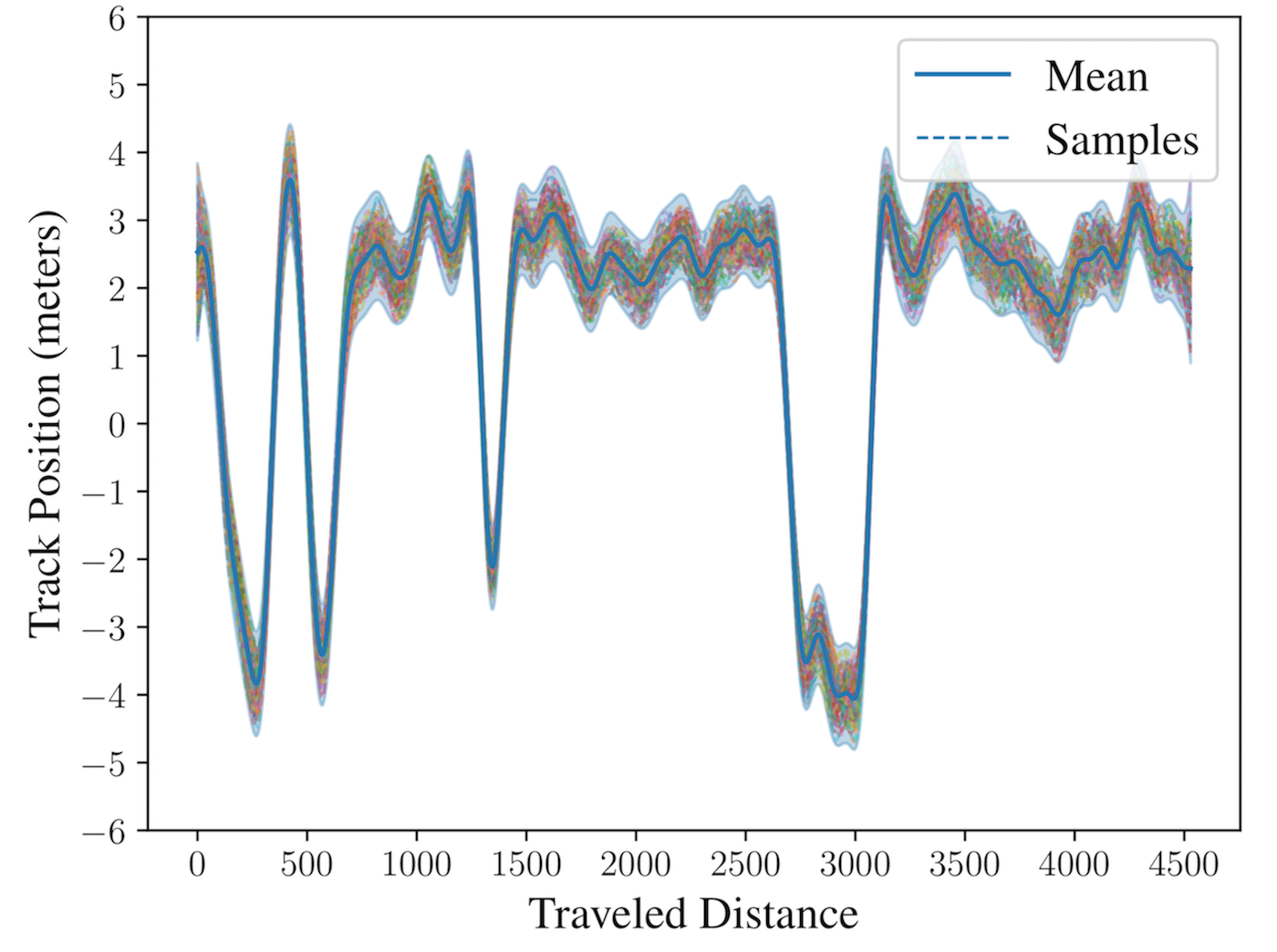}
    \caption*{(a)}
    \end{minipage}
    \vfill
    \begin{minipage}[b]{0.5\textwidth}
   \includegraphics[width=\columnwidth]{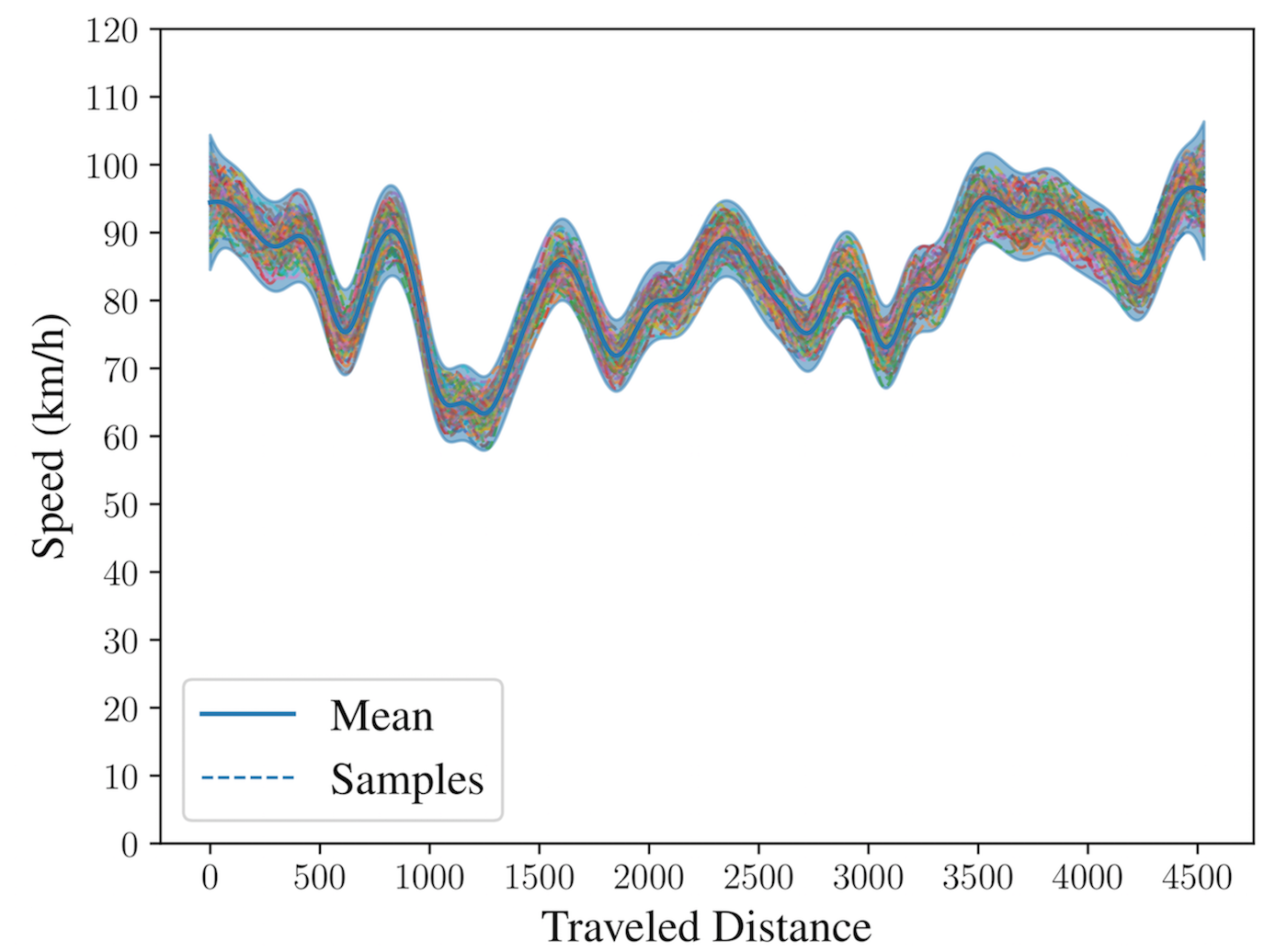}

    \caption*{(b)}
    \end{minipage}
    \caption{Sampling from GPs: 100 random trajectories sampled from the expert driver's distribution of (a) track position (b) speed as a function of arc-length. Each sample is within a 99\% confidence interval. The dashed lines represent the samples, the solid blue lines  the means.}
    \label{fig:bounded_samples_GPs}
\end{figure}
\subsection{Reward Functions}
\label{sec:reward_functions}
In this section, we introduce two reward functions designed to imitate the expert driver's behavior. The first reward function is deterministic and used to reproduce the first moment (mean) of the expert driver's probability distribution for track position and speed. The second one is stochastic and used to recover the first \textit{and} second moments (variance) of the same distributions.

\textit{First moment (mean)}. We consider the following deterministic reward function
\begin{equation}
R(s_t,a_t) = 
\begin{cases}\!
\begin{aligned}[b]
\bigg(\bar{V}_h - |\bar{V}_h - V_a |\bigg)\bigg|_{\sigma(t)} - c_1 {|\bar{D}_h - D_a|}\bigg|_{\sigma(t)} \\
-c_2 |\psi_{t} - \psi_{t-1}| - c_3|\tau_{t}  - \tau_{t-1}|
\end{aligned}
\\
-100\,, \; \mbox{if}\; s_{\sigma(t)} = s_T
\end{cases}\,,
\label{eq:deterministic_reward_function}
\end{equation}
where $s_T$ are termination states defined by 
$s_T$ = \{\textit{obstacle collision, exceeding road boundaries, driving slower than 5 km/h, driving in the opposite direction}\},

and the subscripts \textit{a} and \textit{h} denote agent and human, respectively, and $\sigma(t)$ is the  arc-length at  time $t$. The first term in (\ref{eq:deterministic_reward_function}) measures the deviation of  the mean speed of the agent, $ \bar{V}_a$,  from the mean speed of the expert driver $ \bar{V}_h$. The agent receives higher reward when deviating less from  the expert (positive for $0<V_a< 2\bar{V}_h$ and negative for $V_a > 2\bar{V}_h)$  and a maximal reward of $\bar{V}_h|_{\sigma(t)}$ when $V_a = \bar{V}_h$. The mean speed of the expert driver is obtained from the GP model.
The second term in (\ref{eq:deterministic_reward_function})  defines the difference between the mean track position of the agent and expert driver and $c_1$ is a penalization weight, which was set to 20. The maximum penalty is $-12c_1$, where the value 12 corresponds to two lane widths. The third and fourth term are smoothing terms for steering and torque commands, respectively, and do not incorporate any human data. $c_2$ and $c_3$ are penalization weights that were set to 100 and 10, respectively.

\textit{First and Second Moments (Mean and  variance)}\label{sec:first_and_second_moment}
To recover  mean \textit{and} variance of expert driving behavior, the reward function is defined as follows\footnote{Due to computational limitation, we took a large number of samples (100) that were sufficient to capture the distribution within the range of a 99\% confidence interval (as presented in Section \ref{sec:modeling_human_behavior}). With enough computational power, one can try using on-line sampling.} 
\begin{equation}
R^j(s_t,a_t) = 
\begin{cases}\!
\begin{aligned}[b]
\bigg({V}_h^j - |{V}_h^j - V_a |\bigg)\bigg|_{\sigma(t)} - c_1 {|{D}_h^j - D_a|}\bigg|_{\sigma(t)} \\
-c_2 |\psi_{t} - \psi_{t-1}| - c_3|\tau_{t}  - \tau_{t-1}|
\end{aligned}
\\
-100\,, \; \mbox{if}\; s_{\sigma(t)} = s_T
\end{cases}\,.
\label{eq:stochastic_reward_function}
\end{equation}
The reward function (\ref{eq:stochastic_reward_function}) differs from (\ref{eq:deterministic_reward_function})  by replacing the mean  values  ($\bar{D}_h, \bar{V}_h$) of the expert distributions by samples ($D^j_h, V^j_h$), where the index $j$ denotes sample $j$.  The samples are obtained off-line from the modeled GPs. In each episode, a different sample is randomly chosen to evaluate the reward function. This will induce stochastic behavior in the agent. It is important to emphasize that no underlying dynamic model is assumed here, thus, a model-free framework is applied.

\subsection{Learning Algorithm and  Network Architecture}
\label{sec:learning_algorithem}
We considered two state-of-the-art deep reinforcement learning algorithms to learn a driving policy: Deep Deterministic Policy Gradient (DDPG), \cite{lillicrap2015continuous}, and Proximal Policy Optimization (PPO), \cite{schulman2017proximal}. Ultimately, we decided to use PPO in the learning algorithm. As has already been reported in other studies, \cite{haarnoja2018soft}, hyperparameters tuning in DDPG turns out to be difficult. We observed unstable behavior in spite of extensive hyperparameter exploration (such as different weights initialization, number of hidden layers, number of neurons in each layer) and use of prioritized experience replay (PER), \cite{schaul2015prioritized}. In PPO, we used generalized advantage estimation (GAE) for computing variance-reduced advantage functions. The hyperparameter settings are provided in Table \ref{table:PPO_hyperparameters}.
Our objective is to imitate the stochastic behavior of an expert driver while obeying basic driving rules. 
We extended the vanilla-PPO algorithm by modeling the agent's policy (actor) by a Mixture Density Network with three Gaussians. The number of Gaussians was arbitrarily chosen and no attempt was made to adjust this hyperparameter. We used a different network to approximate the mixing coefficients for each Gaussian. This approach resulted in three different networks -- one for the actor, one for the critic, and one for the mixing coefficients.

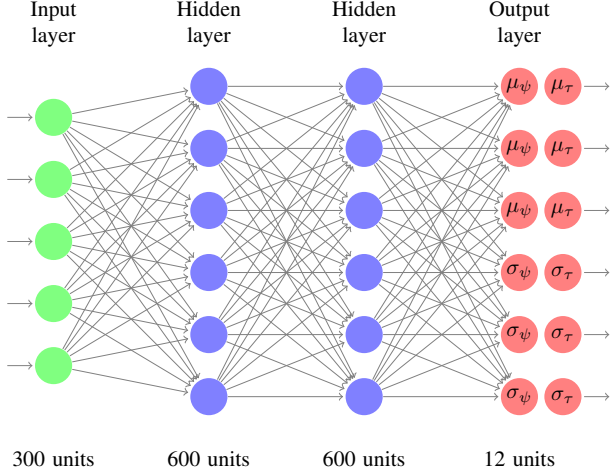
\begin{figure}[ht]
    \resizebox{0.5\textwidth}{!}{
    \centering
    \def\layersep{2.5cm}
\def\halfoutputdim{6}
\def\hiddenlayerdim{6}
\def\inputlayerdim{5}

\begin{tikzpicture}[shorten >=1pt,->,draw=black!50, node distance=\layersep]
    \tikzstyle{every pin edge}=[<-,shorten <=1pt]
    \tikzstyle{neuron}=[circle,fill=black!25,minimum size=17pt,inner sep=0pt]
    \tikzstyle{input neuron}=[neuron, fill=green!50];
    \tikzstyle{output neuron}=[neuron, fill=red!50];
    \tikzstyle{hidden neuron}=[neuron, fill=blue!50];
    \tikzstyle{annot} = [text width=4em, text centered]

    \foreach \name / \y in {1,...,\inputlayerdim}
        \node[input neuron, pin=left:] (I-\name) at (0,-\y) {};

    \foreach \name / \y in {1,...,\hiddenlayerdim}
        \path[yshift=0.5cm]
            node[hidden neuron] (H1-\name) at (\layersep,-\y cm) {};
    
    \foreach \name / \y in {1,...,\hiddenlayerdim}
        \path[yshift=0.5cm]
            node[hidden neuron] (H2-\name) at (\layersep*2,-\y cm) {};

    \foreach \name / \y in {1,...,\halfoutputdim}
        \path[yshift=0.5cm]
            node[output neuron] (O1-\name) at (\layersep*3,-\y cm) {\ifnum\y<4 {$\mu_\psi$} \else {$\sigma_\psi$}\fi};
    \foreach \name / \y in {1,...,\halfoutputdim}
        \path[yshift=0.5cm]
            node[output neuron,pin={[pin edge={->}]right:}] (O2-\name) at (\layersep*3+0.7cm,-\y cm) {\ifnum\y<4 {$\mu_\tau$} \else {$\sigma_\tau$}\fi};;

    \foreach \source in {1,...,\inputlayerdim}
        \foreach \dest in {1,...,\hiddenlayerdim}
            \path (I-\source) edge (H1-\dest);

    \foreach \source in {1,...,\hiddenlayerdim}
        \foreach \dest in {1,...,\hiddenlayerdim}
            \path (H1-\source) edge (H2-\dest);
            
    \foreach \source in {1,...,\hiddenlayerdim}
        \foreach \dest in {1,...,\halfoutputdim}
            \path (H2-\source) edge (O1-\dest);

    \node[annot,above of=H1-1, node distance=1cm] (h1-1) {Hidden layer};
    \node[annot,left of=h1-1] {Input layer};
    \node[annot,right of=h1-1](h2-1) {Hidden layer};
    \node[annot,right of=h2-1] {Output layer};
    
    \node[annot,below of=H1-6, node distance=1cm] (h1-6) {600 units};
    \node[annot,left of=h1-6] {300 units};
    \node[annot,right of=h1-6](h2-6) {600 units};
    \node[annot,right of=h2-6] {12 units};

\end{tikzpicture}}
    \caption{The actor-network architecture: Input with 300 units, two hidden layers with 600 units each, and an output layer with 12 units. The output layer is divided into blocks, corresponding to the number of Gaussians in the mixture. The output nodes provide the means ($\mu_\psi$ and $\mu_\tau$) and  variances ($\sigma_\psi$ and $\sigma_\tau$) for steering and the throttle commands, respectively.}
    \label{fig:actor_net_arch}
\end{figure}

\textit{Actor Network Architecture}: The actor-network consisted of an input layer of 300 units, 2 hidden layers of 600 units each and an output layer of 12 units. We used ReLU activation functions between the different layers except for the output layer, where two different activation functions were used to bound the mean and variances. The 12 output units were divided into 2, where the first 6 units represent the mean values for for steering and throttle commands, $\mu_\psi$ and $\mu_\tau$, and the other 6 units represent the variances, $\sigma_\psi$ and $\sigma_\tau$, for the three Gaussians in the mixture. The first 6 values (of the means) are passed through a SoftSign activation function in order to bound the values between -1 to 1, as the control commands are bounded. The last 6 values (of the covariances) are passed through a customized activation function
of the form $f(x) =  1/16\cdot \mbox{Sigmoid}(3x)$. This form  was chosen
in order to have a non-negative variance for each Gaussian and to limit the variance to a maximal value. We noticed that larger values in the variance led to a disproportionate amount of noise in the policy and caused the agent to get stuck in local minima at the very first start.  The actor-network architecture  is illustrated in Figure \ref{fig:actor_net_arch}.

\textit{Critic Network Architecture}: The critic has the same network architecture as the actor except for the output layer, which consists of only one neuron output, the Q-value. A ReLU activation function was used in all layers except for the last layer, where  a linear activation function was used.\\
\indent
\textit{Mixing Coefficients Network Architecture}: Similarly, the mixing coefficient network\footnote{While the mixing coefficients could be included as part of the actor's output layer, no such attempt was made.} has the same architecture as the actor and critic except that the output layer consists of 3 neurons (the number of Gaussians in the mixture). ReLU activation function were used for all layers except for the output layer, for which a \textit{softmax} activation function was applied in order to maintain the constraint $\sum \limits_{i} \alpha_i = 1$.

\subsection{Exploration Strategy and Policy Update}

One of the main challenges in RL is the problem of exploration. PPO is an on-policy algorithm that trains a stochastic (Gaussian) policy and exploration occurs by sampling actions from this policy.

Thus, the amount of exploration is affected by the value of the covariance matrix and the training procedure. In  standard training procedures  fixed initial states and fixed episode lengths are used, which often result in unsuccessful learning. To improve exploration and avoid early termination, we used \textit{reference state initialization} as suggested in \cite{peng2018deepmimic}. We initialized the speed by sampling from a uniform distribution between 30 to 90 km/h.  High variability in the policy at the beginning of training caused the agent to terminate after a few number of steps (30-40). A full round of the track required about 2000 steps. To improve learning
we implemented a dynamic batch size that grows with the agent's performance. The pseudocode for dynamic batch size update is described in the Algorithm\ref{algorithem:dynamic_batch_size_update}.

\begin{algorithm}[h]
\caption{Dynamic Batch Size Update}\label{algorithem:dynamic_batch_size_update}
\begin{algorithmic}[1] 
\STATE $\text{initialize batch size \textit{B}, and mini batch size \textit{MB}}$
\STATE $\text{initialize \textit{max\_length}} \leftarrow 0$
\STATE $\text{initialize Memory}$
\WHILE{$\text{not done}$}
    \STATE \texttt{<Get Initial State>}
    \FOR{$step = 1,\dots,\text{max episode steps}$}
        \STATE \texttt{<Run The Algorithm>}
        \STATE \texttt{<Store (s,a,r,s') Into Memory>}
        \IF{ $\textit{step} \mod \textit{B}=0$ \OR $\textit{ completed\_rounds}$ }
            \STATE \texttt{<Update The Policy>}
            \STATE \texttt{<Clear The Memory>}
            \STATE \texttt{<Reset Environment>}
             \STATE \texttt{<Break For Loop>}
        \ENDIF
        \IF{$\textit{step}>\textit{max\_length}$ }
            \STATE $\textit{max\_length} \leftarrow \textit{step}$
            \STATE $\textit{rem} \leftarrow  2\cdot \textit{max\_length} \!\!\!\mod \textit{MB}$
            \STATE $B \leftarrow \max(B, 2\cdot \textit{max\_length} - \textit{rem})$
        \ENDIF
    \ENDFOR
\ENDWHILE

\end{algorithmic}

\end{algorithm}

\noindent

The (Boolean) parameter $\textit{completed\_rounds}$  indicates whether the agent completed two rounds or not. The parameter $\textit{max\_length}$ is the maximum number of steps the agent has taken in a row from the start of an episode to its end. The $\textit{rem}$ parameter is the number of steps remaining to reach twice\footnote{A multiplier of $2$ is used, so the batch size is about twice the maximum number of times the agent can execute per episode.} the $\textit{max\_length}$. The $\textit{rem}$ guarantees that the new batch ($B$) can be divided into equal mini-batches ($MB$). $B$ and $MB$ were initialized to 512 and 256, respectively.

In addition, an entropy bonus can be added to the loss function in PPO to motivate more exploration \cite{williams1992simple, mnih2016asynchronous}. Here, we used an entropy coefficient of 0.005 to ensure sufficient exploration, as suggested in past works.

\section{RESULTS}
\begin{figure}[ht]
    \centering
    \begin{minipage}[b]{0.49\textwidth}
     \includegraphics[width=\columnwidth]{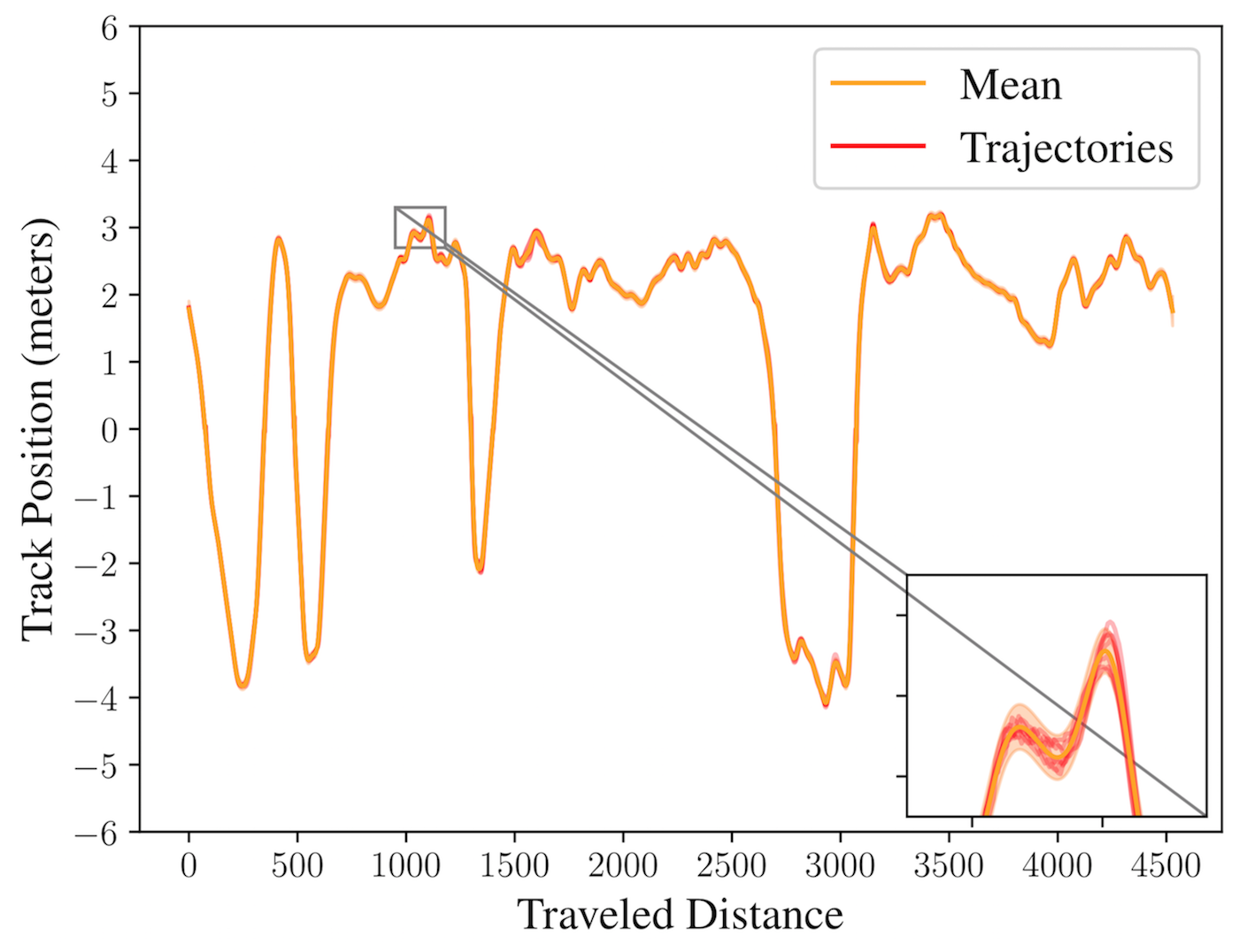}
    \caption*{(a)}
    \end{minipage}
    \vfill
    \begin{minipage}[b]{0.49\textwidth}
    \includegraphics[width=\columnwidth]{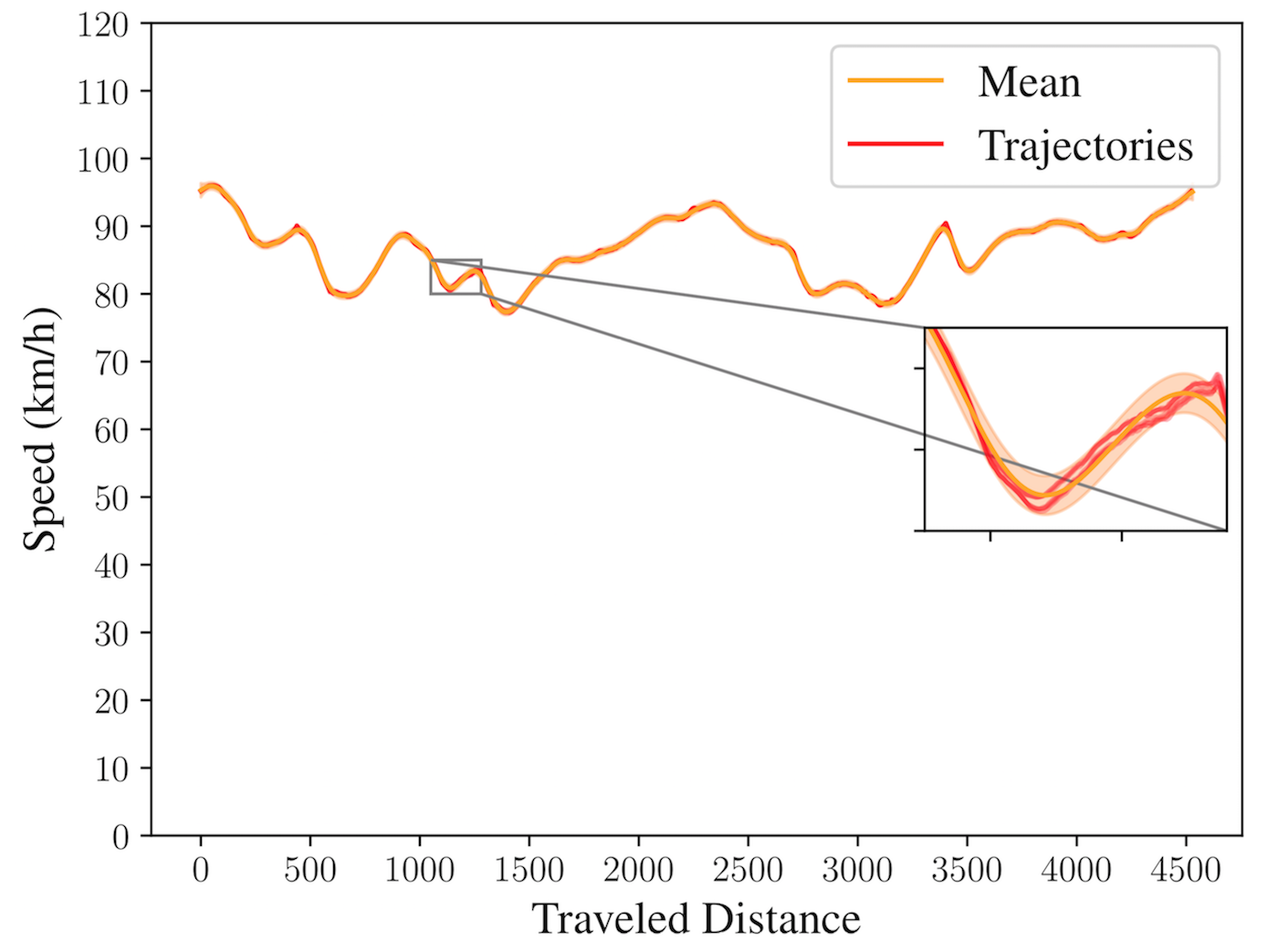}
    \caption*{(b)}
    \end{minipage}
    \caption{Agent driving and GP regression using a \textit{deterministic} reward function: (a) Track position. (b) Speed. Trajectories are shown in red from 7 rounds of  driving. The mean of the GP is shown as an orange line and the 99\% confidence interval as an orange shaded area (insert).}
    \label{fig:agent_deterministic_vs_human_dist}
\end{figure}

\begin{figure}[ht]
    \centering
    \begin{minipage}[b]{0.49\textwidth}
    \includegraphics[width=\columnwidth]{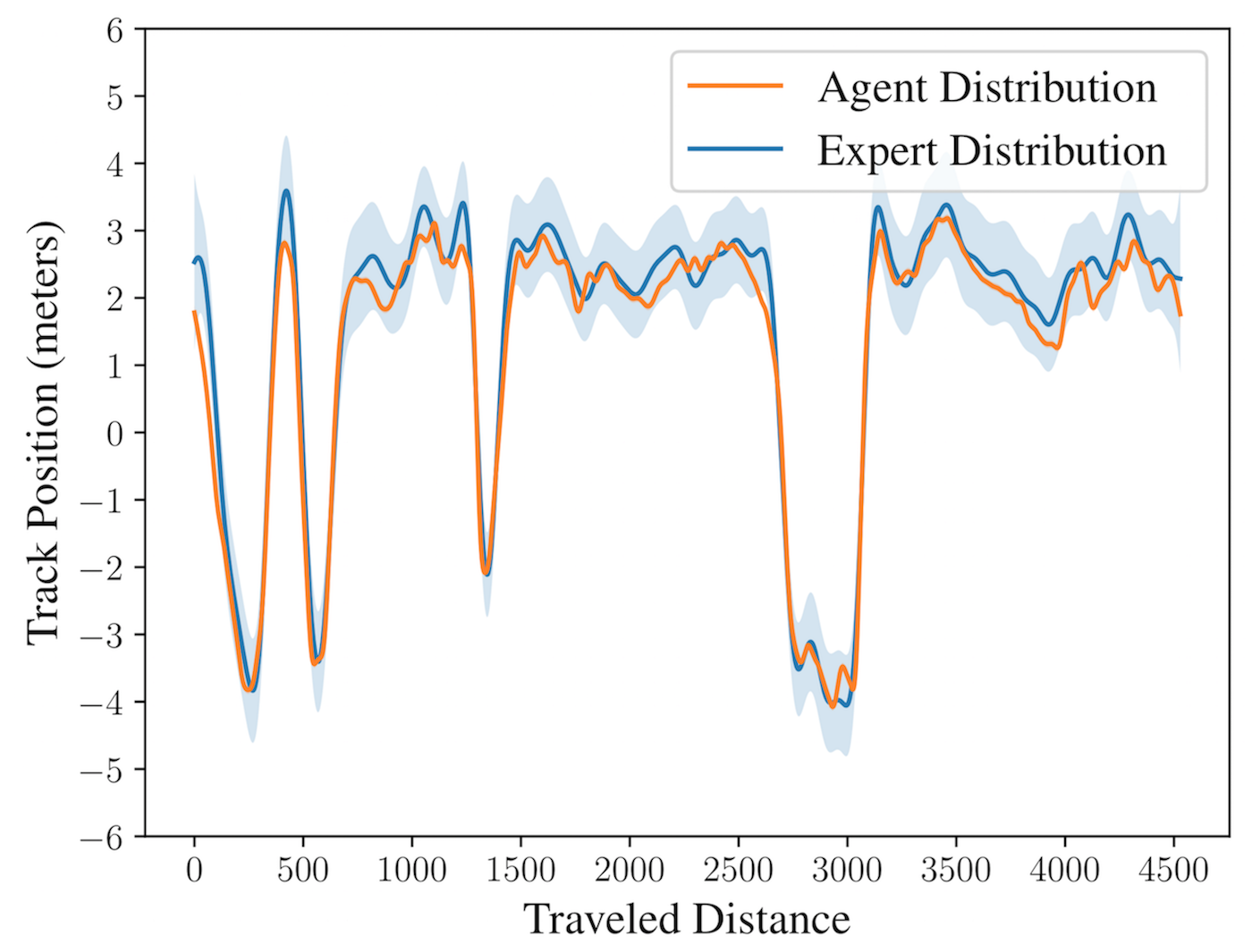}
    \caption*{(a)}
    \end{minipage}
    \vfill
    \begin{minipage}[b]{0.49\textwidth}
    \includegraphics[width=\columnwidth]{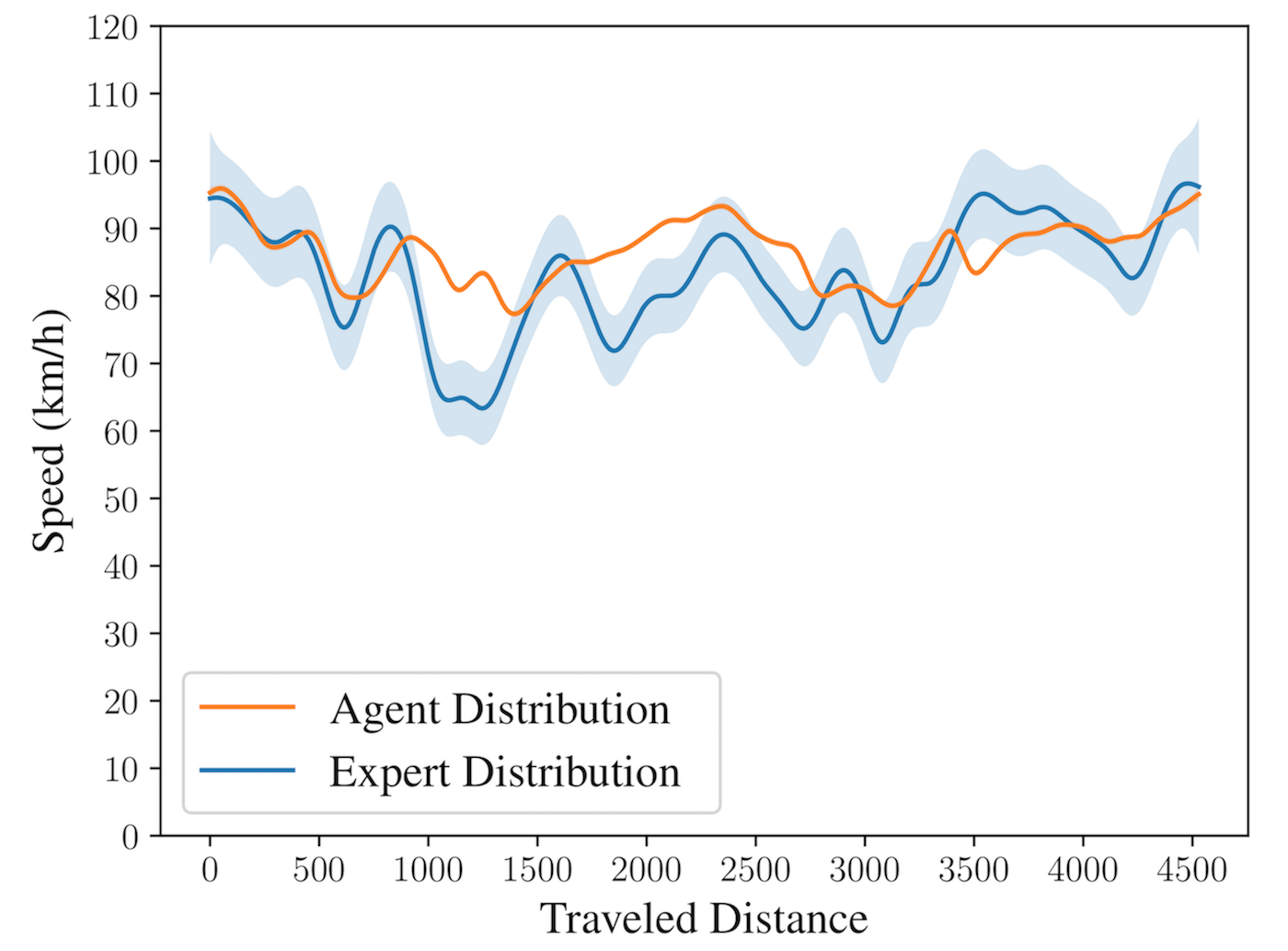}
    \caption*{(b)}
    \end{minipage}
    \caption{Comparison of agent and human expert driving distributions using a \textit{deterministic} reward function: (a) Track position (b) Speed. The means of the GPs are shown as solid lines and the 99\% confidence intervals as shaded areas (insert).}
    \label{fig:agent_deterministic_policy_results}
\end{figure}

We studied imitation of an expert driver in an obstacle avoidance task by applying a model-free deep reinforcement learning algorithm to a learning agent. We first present our results for recovering the first moment (mean) of the expert distribution by using a \textit{deterministic} reward function, followed by the results for recovering mean \textit{and} variance of the expert driver distribution by using a \textit{stochastic} reward function.\\
\textit{Trained Agent Policy -- Mean:} \label{sec:recover_mean_behavior}
We first let the agent drive in a deterministic mode, in which only mean values of the control output from the actor-network are considered while the variance is ignored. Figure \ref{fig:agent_deterministic_vs_human_dist} shows trajectories (red) for track position and speed from seven rounds of agent driving agent after training. GPs were fitted to the data. Shown is the mean (solid lines) and a 99\% confidence interval (shaded area). As can be seen, the agent converged to a deterministic behavior with minimal fluctuations.
Figure \ref{fig:agent_deterministic_policy_results} shows the comparison between agent driving and human expert driving.
The agent recovered track position well, but tends to overestimate speed. We first hypothesized that the latter may be caused by the  presence of local minima. However, an increase of training time (up to 5 days) did not lead to  any changes although a recovery  from local minima should have been possible. A second explanation for the mismatch may be the limited expressiveness of  the actor policy (one Gaussian), but as shown below, a Gaussian mixture model for the policy led to similar results.  Finally, we concluded that the discrepancy results from an agent acting in a partially observable environment as elaborated below.\\

\begin{figure}[ht]
    \centering
    \begin{minipage}[b]{0.49\textwidth}
	\includegraphics[width=\textwidth]{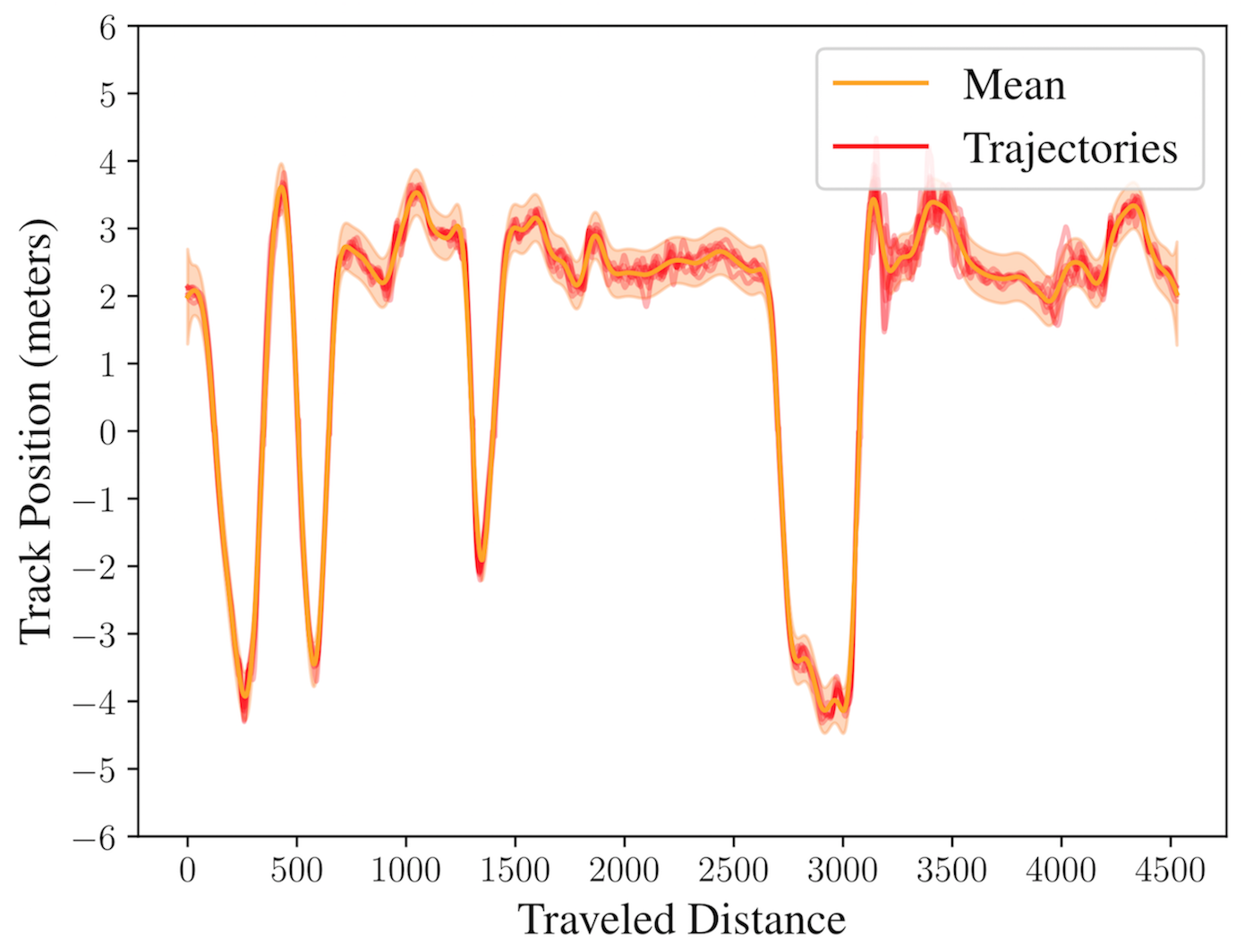}
    \caption*{(a)}
    \end{minipage}
    \vfill
    \begin{minipage}[b]{0.49\textwidth}
	\includegraphics[width=\textwidth]{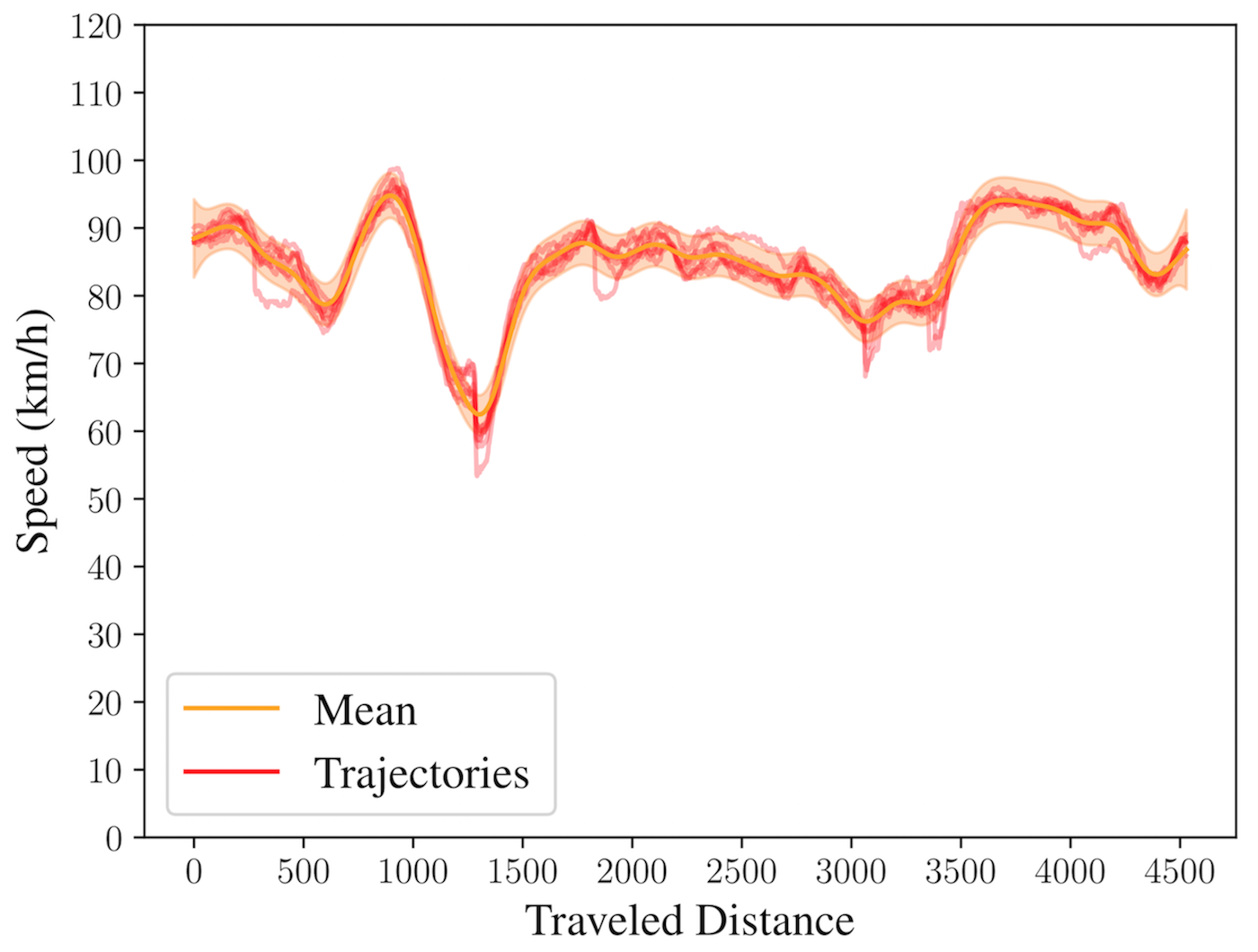}
    \caption*{(b)}
    \end{minipage}
    \caption{Agent driving and GP regression using a \textit{stochastic} reward function: (a) Track position. (b) Speed.
     Trajectories of seven rounds of driving are shown in red. The mean of the GP is shown as a solid line (orange) and a 99\% confidence interval as an orange shaded area.}
    \label{fig:agent_policy_results}
\end{figure}

\begin{figure}[ht]
    \centering
    \begin{minipage}[b]{0.49\textwidth}
    	\includegraphics[width=\textwidth]{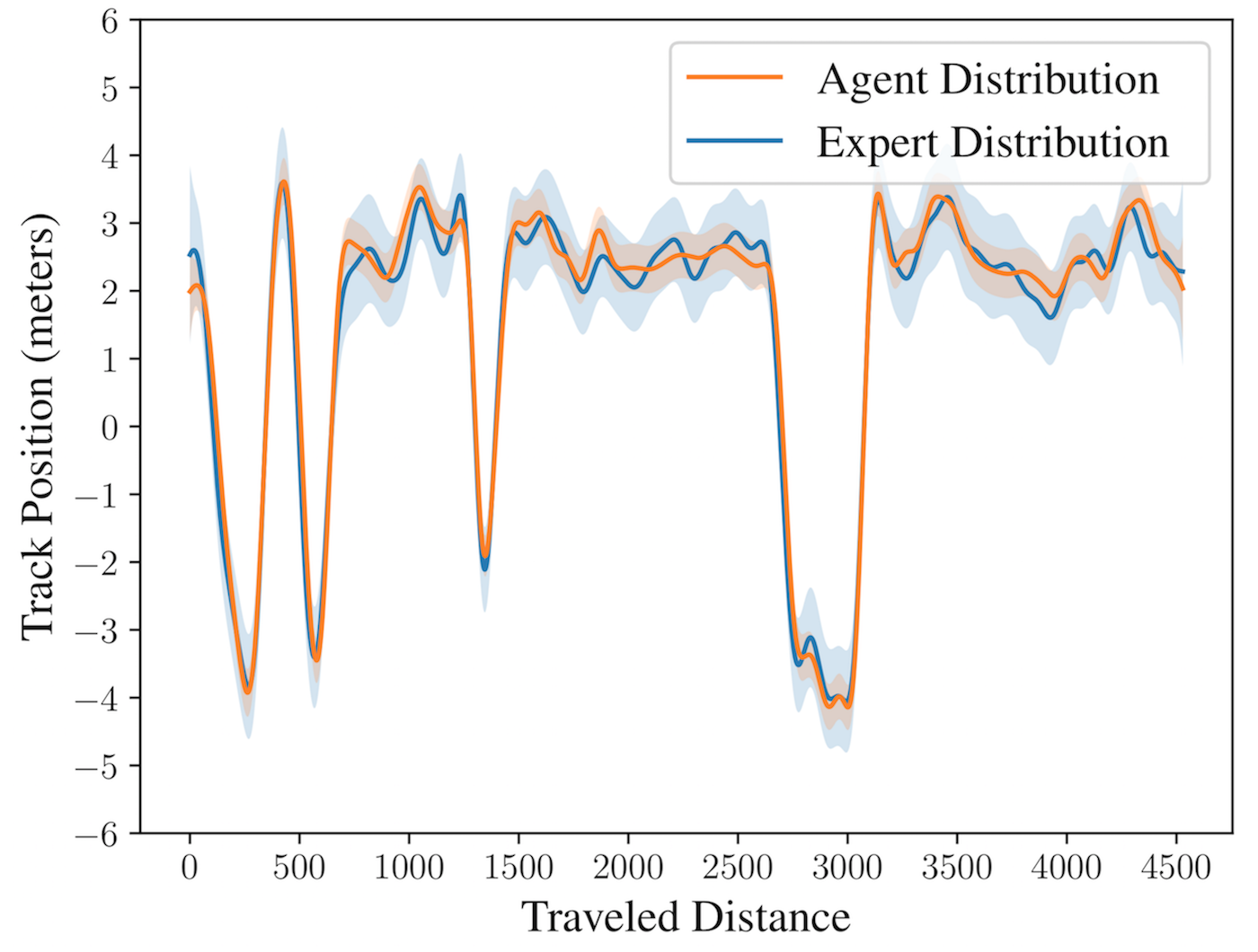}
    \caption*{(a)}
    \end{minipage}
    \vfill
    \begin{minipage}[b]{0.49\textwidth}
    \includegraphics[width=\textwidth]{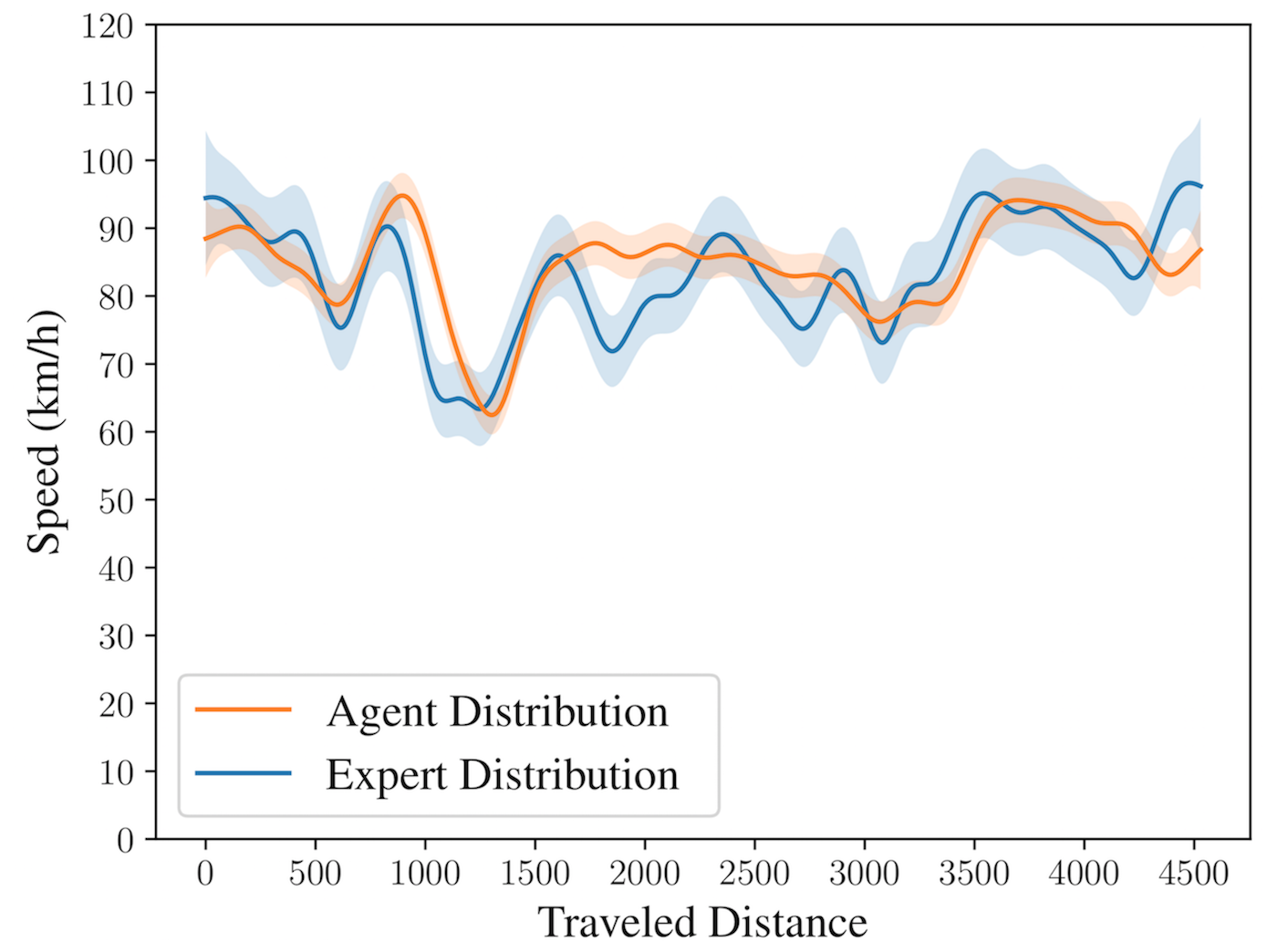}
    \caption*{(b)}
    \end{minipage}
    \caption{Comparison of agent and expert driving distributions using a \textit{stochastic} reward function:  (a) Track position. (b) Speed. 
    The means of the modeled GPs are shown as solid lines and a 99\% confidence interval is indicated as shaded areas.}
    \label{fig:agent_vs_human_dist}
\end{figure}

\textit{Trained Agent Policy -- Mean and Variance:}
\label{sec:recover_mean_and_variability}
To reproduce  the variability of expert human driving, a stochastic reward function, as given in (\ref{eq:stochastic_reward_function}), was used. Figure \ref{fig:agent_policy_results} shows trajectories of agent driving after training and GP regression for the two measurement variables.
Figure \ref{fig:agent_vs_human_dist} shows the comparison of agent and human expert driving.
As before, the agent overestimated speed, but the actor is now modeled by a MDN, which results in an increased expressiveness of the policy  and a better reproduction of the expert distribution.
 We believe that the remaining difference in speed estimation results from an agent acting in a partially observable environment. In particular, the agent's limited field of view (set to 300 m) has a significant impact on speed control.

For instance, consider the two scenarios of Figure \ref{fig:speed_limiting_case}a and \ref{fig:speed_limiting_case}b, where a car is approaching two similar curved road segments. Within the agent's field view (indicated by the back circle) the two states are identical. The human driver, on the other hand, can easily initiate a quick glance towards the obstacle behind the curve in {\ref{fig:speed_limiting_case}a, and thus, considers these states as different. As long as we do not provide the agent with additional information to distinguish these two states, it will probably learn to assign the same speeds.

\begin{figure}[ht]
    \centering
    \begin{minipage}[b]{0.49\textwidth}
    \includegraphics[width=\textwidth]{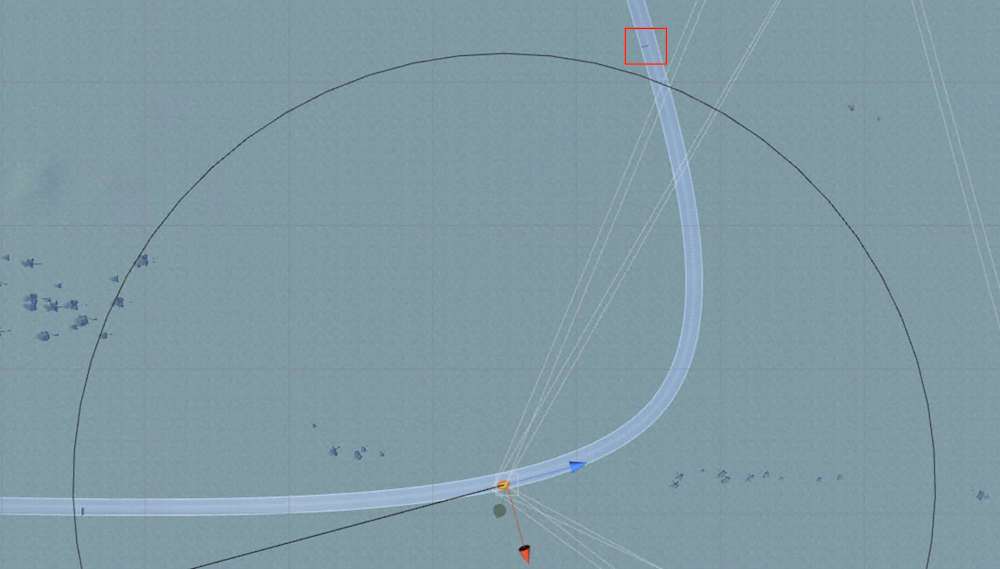}
    \caption*{(a)}
    \end{minipage}
    \vfill
    \begin{minipage}[b]{0.49\textwidth}
    \includegraphics[width=\textwidth]{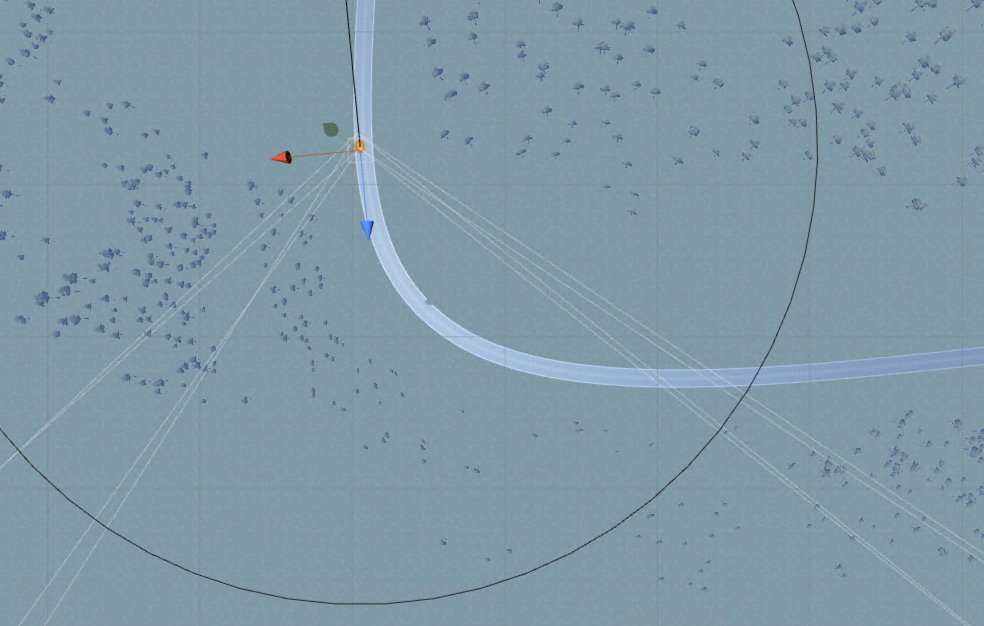}
    \caption*{(b)}
    \end{minipage}
    \caption{
  Speed control in a partially observable driving environment:  (a) Road segment with obstacle (red square) beyond the agent's field of view (black circle) (b) Similar road segment to (a) but without obstacle. Within the agent's field of view there are no obstacles. The blue arrow shows the car traveling direction.}
    \label{fig:speed_limiting_case}
\end{figure}

\textit{Turing Test}: To assess the similarity between human and machine driving we have proposed to compare distributions of kinematic features. In addition, one may use a Turing test to ultimately decide whether a machine has human-like driving skills. Such a test would consist of a human passenger that is placed in the back seat of a vehicle  without seeing the driver cabin and being queried whether the car is driven by a machine or human. If the passenger cannot decide who is driving, the Turing test is passed. As we could not execute a Turing test in a real environment, we performed a visual Turing test at the Israeli Conference on Robotics (ICR) 2019, where we have presented parts of this work. We showed the audience a video (\href{https://youtu.be/rk6M0-5UR4Y}{\textit{Video 1}}) of recorded human expert driving and our machine driving agent. Interestingly, the audience had difficulties to identify the human driver (about 40\% correct, 60\% incorrect) indicating that the agent has acquired human-like driving skills for this task.

\textit{Generalization}: 
We have performed initial tests of generalization by analyzing the agents' behavior on new road tracks for different obstacle distributions. We first tested 
whether an agent based on a deterministic or stochastic reward function can generalize better to a new road with a different obstacle distribution. As expected, we found that the latter showed better generalization capabilities, being able to avoid all obstacles (\href{https://youtu.be/0S8ca4NuSck}{\textit{Video 2}}). 
The agent using a stochastic reward was then tested on a new road (total length 3.14 km) with curvy and straight sections and three different obstacle distributions: (i) alternating obstacle locations with a fixed distance of  50 m, (ii) randomized obstacle loactions with inter-obstacle distances sampled from a Gaussian distribution (mean: 100 m, std: 10 m) and  lane locations (left or right) from a binary random variable. (iii) same as in (ii) but instead of placing one obstacle, a batch of obstacles (2-4) was placed by sampling from a discrete uniform distribution (\href{https://youtu.be/e79EKR_IgTI}{\textit{Video 3}}). Although the agent generated rather human-like driving skills, we observed that the agent failed to avoid obstacles, which were placed in road sections with high curvature, a combination not previously seen in the training phase.

\section{CONCLUSIONS}
In this paper we have introduced a novel and easy-to-implement model-free approach to imitate the behavior (mean and variability) of an expert driver. For this purpose, we have built a full driving simulator environment in Unity and focused on rather simple obstacle avoidance tasks, in which static obstacles were distributed arbitrarily on a two-lane highway road.

We have shown that our method can reproduce well human expert behavior in an environment with high dimensional state space. Track position was recovered better than speed and we concluded that the latter is related to an agent acting in a partially observable environment. First generalization tests on new road tracks with different obstacle distributions showed that the agent  developed human-like obstacle avoidance skills except in cases that deviated significantly from what has been observed during training.

This work has several limitations. 
First, quantifying generalization in RL is a challenging problem. Cobbe et al, \cite{cobbe2018quantifying}, have introduced a generalization metric and have shown that a large number of training environments are required for good generalization. Second, we have not investigated different state-space representations, such as the inclusion of camera images, neither have we studied different network architectures, such as RNNs, which may represent temporal correlations between states better and lead to improved results. 
Finally, we have not addressed the problem of moving obstacles as given by other vehicles or motorcycles. These extensions need to be considered in future work.

\section*{APPENDIX}

\subsection{Hyperparameters}

\begin{table}[h]
\begin{center}
\begin{tabular}{ l|l }
Hyperparameter & Value \\
\hline
Minibatch size (MB)             & 1024             \\
Adam stepsize                   & $1\times10^{-1}$ \\
Num. epochs                     & 5                \\
Discount $(\gamma)$             & 0.98             \\
GAE parameter $(\gamma)$        & 0.95             \\
Number of actors                & 1                \\
Number of Gaussians             & 3                \\
Clipping parameter $(\epsilon)$ & 0.2              \\
MSE loss coeff. $c_1$           & 0.5              \\
Entropy coeff. $c_2$            & 0.005            \\
\end{tabular}
\caption{PPO hyperparameters used in the experiment.}
\label{table:PPO_hyperparameters}
\end{center}
\end{table}

\section*{ACKNOWLEDGMENT}

The authors would like to thank Ron Pick for his helpful advice regarding the client-server protocol of the simulator environment, and Rotem Duffney for helping to record the videos.

\bibliographystyle{ieeetr}

\bibliography{literature}

\end{document}